\documentclass[letterpaper, 10 pt, conference]{ieeeconf}  

\IEEEoverridecommandlockouts                              
\usepackage{array}
\usepackage{tabu}
\usepackage{graphicx}
\usepackage{multirow}
\usepackage{subcaption}
\usepackage{fancyhdr}

\fancypagestyle{firstpage}{%
  \rhead{ICARCV 2018}
  \lfoot{\small{\copyright2018 IEEE. Final paper is published at: https://ieeexplore.ieee.org/document/8581283. Personal use of this material is permitted. Permission from IEEE must be obtained for all other users, including reprinting/republishing this material for advertising or promotional purposes, creating new collective works for resale or redistribution to servers or lists, or reuse of any copyrighted components of this work in other works.}}
}

\title{\LARGE \bf A Multi-Face Challenging Dataset for Robust Face Recognition}

\author{Shiv Ram Dubey and Snehasis Mukherjee
\thanks{The authors are with the Computer Vision Group, Indian Institute of Information Technology, Sri City, Andhra Pradesh-517646, India. Email: \{srdubey, snehasis.mujherjee\}@iiits.in.}
}

\begin{document}

\maketitle
\thispagestyle{firstpage}

\begin{abstract}
Face recognition in images is an active area of interest among the computer vision researchers. However, recognizing human face in an unconstrained environment, is a relatively less-explored area of research. Multiple face recognition in unconstrained environment is a challenging task, due to the variation of view-point, scale, pose, illumination and expression of the face images. Partial occlusion of faces makes the recognition task even more challenging. The contribution of this paper is two-folds: introducing a challenging multi-face dataset (i.e., IIITS\_MFace Dataset) for face recognition in unconstrained environment and evaluating the performance of state-of-the-art hand-designed and deep learning based face descriptors on the dataset. The proposed IIITS\_MFace dataset contains faces with challenges like pose variation, occlusion, mask, spectacle, expressions, change of illumination, etc. We experiment with several state-of-the-art face descriptors, including recent deep learning based face descriptors like VGGFace, and compare with the existing benchmark face datasets. Results of the experiments clearly show that the difficulty level of the proposed dataset is much higher compared to the benchmark datasets.
\end{abstract}



\section{Introduction}

Face detection and recognition from still images is an active research area in computer vision \cite{pietikainen}. Most of the state-of-the-art face recognition approaches were restricted to the controlled environments such as frontal pose \cite{bereta}. Detailed survey on face recognition tasks have been conducted many a time by the researchers \cite{smeulders,huang,yang}. 

Recently, recognizing faces in the wild images, has become an emerging area of research in computer vision \cite{lfw}. Face recognition in unconstrained environment is still an unsolved problem due to the various levels of challenges like part or full occlusion of faces, varying illumination, multiple posture of faces, expressions on faces, etc. The face recognition task becomes even harder when multiple such challenges are present simultaneously. In order to facilitate the face detection and recognition research, we propose a multi-face challenging dataset including all such challenges discussed above. This dataset will be publicly available to the research community.

\begin{figure}[!t]
  \includegraphics[width=\linewidth, height=6cm]{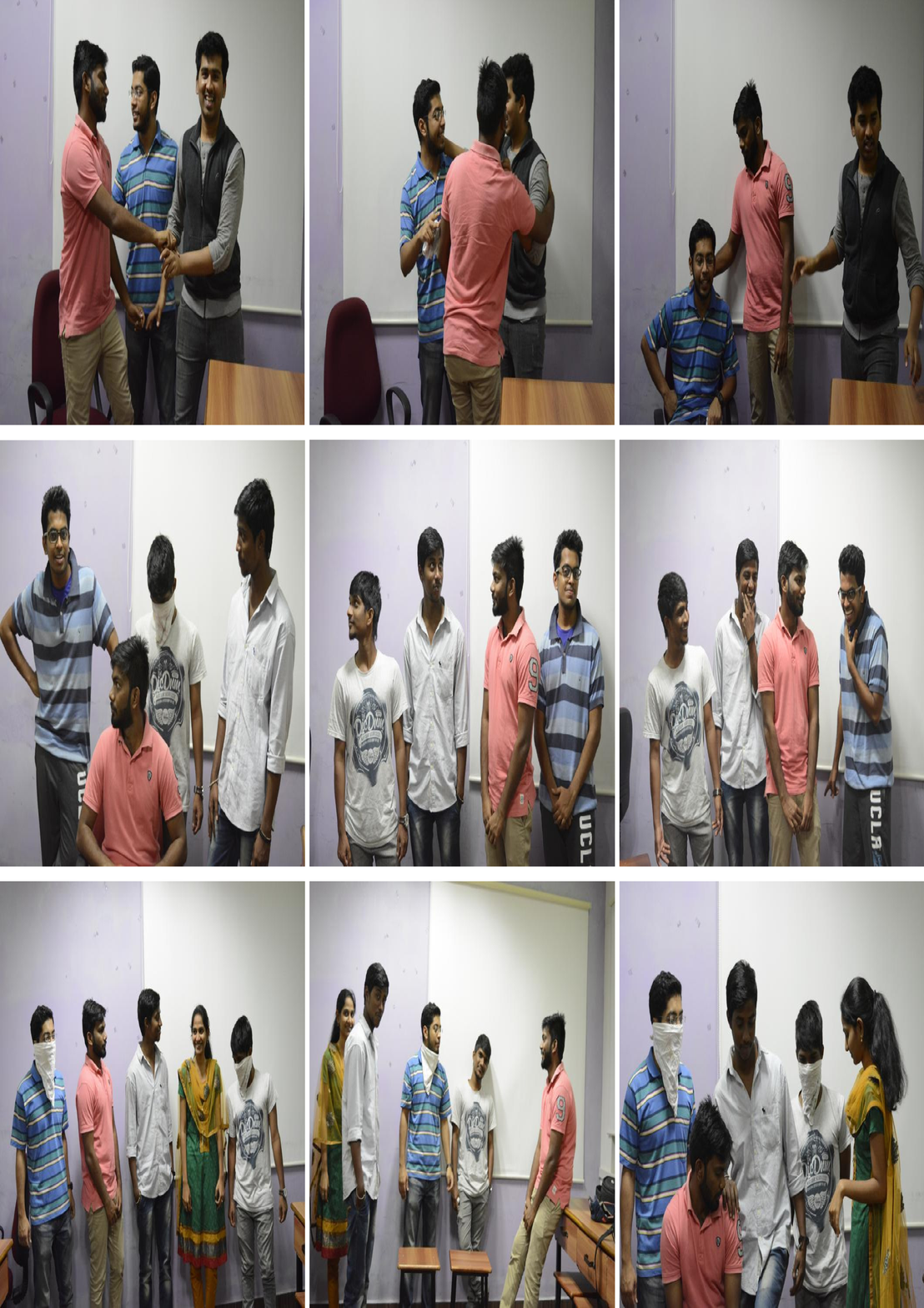}
  \caption{Sample images from original gallery set of the proposed IIITS\_MFace dataset. The various challenges like pose variation, occlusion, illumination changes, orientations, etc. can be observed.}
  \label{fig:1}
\end{figure}

A few publicly available datasets exist in the literature for face recognition and detection, involving challenges like different side poses, occluded faces, varying light intensities,etc. For instance, the AT \& T face database \cite{attdb} has only grayscale and frontal face images. The AR face database \cite{ardb} contains faces with different facial expressions, varying illumination, and occlusions in the face images. This database is having only single face images with uniform background. The CroppedYale dataset contains faces only with the illumination variations \cite{CroppedYale}. The LFW face dataset is challenging and captured under unconstrained environment \cite{lfwdb} with single face images. For the comparison purpose, we have used CroppedLFW version of this dataset \cite{CroppedLFW}. The PaSC face dataset consists of pose, illumination and blur effects \cite{pasc}. Total 8718 faces from 293 subjects are present after applying the Viola Jones object detection method \cite{viola} for face localization in PaSC dataset. The PubFig dataset is another challenging dataset consisting  of images in unconstrained environment \cite{pubfig}. Variations in lighting, expression and pose effects are present in the PubFig dataset with total 6472 images from 60 individuals. The dead urls are removed while downloading the PubFig images. However, none of the existing face datasets offer multiple faces in the images. The proposed IIITS\_MFace is a new dataset for face recognition in images containing multiple faces. Moreover, in addition to the various challenges involved in the state-of-the-art datasets, the images of the proposed dataset are captured in uneven and varying background, which was missing in state-of-the-art datasets. Some sample images from the proposed dataset are shown in Figure \ref{fig:1}.

We show the complicacy in the proposed IIITS\_MFace dataset, by applying state-of-the-art hand-designed as well as deep learning based face recognition techniques on the dataset. We emphasize on local image descriptors, considering the recent success of local image descriptors on face recognition task. Several efforts have been made to apply local image descriptors for face recognition. Local Binary Pattern (LBP) is proposed by Ahonen et al. for the face representation \cite{lbp}. LBP is computed by finding a binary pattern of 1 and 0 for each neighbor of a center pixel. The bit is coded as 1 if the intensity value of neighbor is greater than or equal to the intensity value of center pixel; otherwise it is coded as 0. Local Ternary Pattern (LTP) is the extension of LBP by introducing two thresholds for uniform illumination robust face recognition \cite{ltp}. The LBP over four derivative images corresponding to four directions are computed and concatenated to form the Local Derivative Pattern (LDP) \cite{ldp}. The concept of high order directional gradient is used to find the Local Directional Gradient Pattern (LDGP) to extract the local information of the image \cite{ldgp}. In the recent advancements, Semi-structure Local Binary Pattern (SLBP) \cite{slbp}, Local Vector Pattern (LVP) \cite{lvp}, and Local Gradient Hexa Pattern (LGHP) \cite{lghp} descriptors are proposed for the unconstrained face recognition. The VGGFace CNN descriptor \cite{vggface} is very discriminative and based on the deep learning technique. We experimented with all these descriptors on the proposed dataset. Next we provide a detailed description of the proposed dataset.

\section{Proposed IIITS\_MFace Dataset}
The images in the proposed IIITS\_MFace dataset are captured by cameras of multiple mobile phones to make it more realistic with respect to the real world face recognition problem. A lot of variations in terms of pose, masked, spectacles, number of subjects, illumination, occlusion, etc. are present in the dataset to make it as unconstrained as possible. The proposed dataset is divided into two sections with seven subjects including six male and one female. The two sections of the proposed dataset are named as Gallery Set and Probe Set. The IIITS\_MFace dataset is publicly available for research purpose only\footnote{https://sites.google.com/a/iiits.in/snehasis-mukherjee/datasets-1}.

\begin{table}[!t]
\caption{A summary of gallery set in terms of the variations  like Frontal/Non-frontal pose and Masked/Unmasked}
\label{t1}
\centering
\newcolumntype{D}{>{\small\centering}p{0.09\linewidth}}
\newcolumntype{E}{>{\small\centering}p{0.115\linewidth}}
\newcolumntype{F}{>{\small\centering}p{0.18\linewidth}}
\begin{tabular} {|D|E|E|E|F|D|}
\hline 
Subject ID & \#Frontal Masked & \#Frontal Unmasked & \#Non-frontal Masked & \#Non-frontal Unmasked & \#Total Faces \tabularnewline
\hline
1 & 0 & 27 & 0 & 67 & 94 \tabularnewline
2 & 0 & 59 & 0 & 121 & 180 \tabularnewline
3 & 27 & 31 & 47 & 54 & 159 \tabularnewline
4 & 0 & 4 & 0 & 13 & 17 \tabularnewline
5 & 7 & 17 & 20 & 39 & 83 \tabularnewline
6 & 0 & 18 & 0 & 68 & 86 \tabularnewline
7 & 1 & 29 & 6 & 33 & 69 \tabularnewline
\hline 
Total & 35 & 185 & 73 & 395 & 688\tabularnewline
\hline
\end{tabular}
\end{table}

\begin{table}[!t]
\caption{A summary of gallery set in terms of the \#faces with and without spectacles}
\label{t2}
\centering
\newcolumntype{D}{>{\small\centering}p{0.16\linewidth}}
\newcolumntype{E}{>{\small\centering}p{0.23\linewidth}}
\begin{tabular}{|D|E|E|D|}
\hline Subject ID & \#With Spectacles & \#Without Spectacles & \#Total Faces\tabularnewline
\hline
1 & 0 & 94 & 94 \tabularnewline
2 & 0 & 180 & 180 \tabularnewline
3 & 159 & 0 & 159 \tabularnewline
4 & 17 & 0 & 17 \tabularnewline
5 & 0 & 83 & 83 \tabularnewline
6 & 0 & 86 & 86 \tabularnewline
7 & 0 & 69 & 69 \tabularnewline
\hline Total & 176 & 512 & 688\tabularnewline
\hline
\end{tabular}
\end{table}

\begin{figure*}[!t]
\centering
  \includegraphics[width=.97\linewidth, height=11cm]{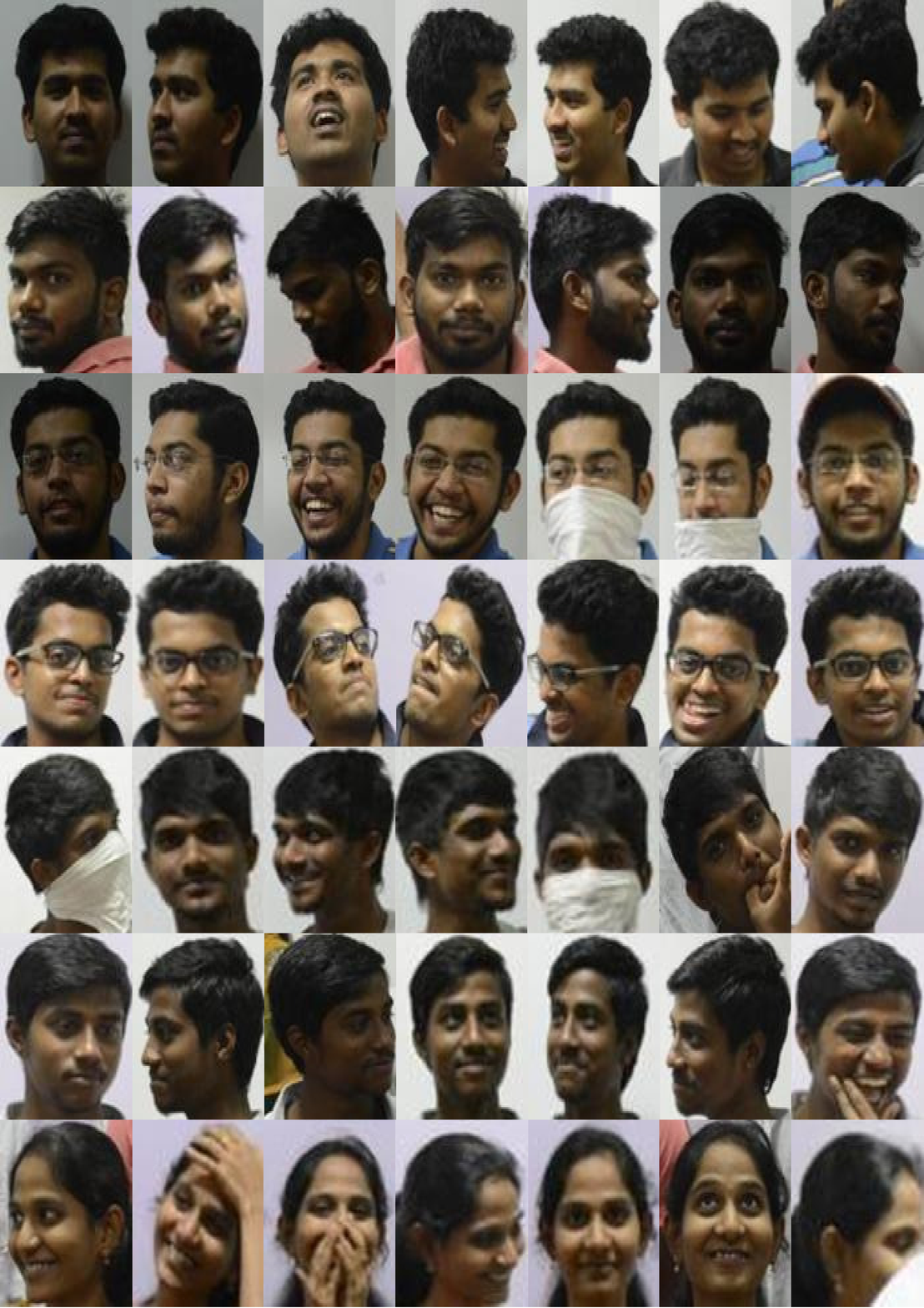}
  \caption{Sample seven faces per subject from gallery set. Each row corresponds to a subject.}
  \label{fig:2}
\end{figure*}

\begin{table}[!t]
\caption{A summary of probe set in terms of the variations  like Frontal/Non-frontal pose and Masked/Unmasked}
\label{t3}
\centering
\newcolumntype{D}{>{\small\centering}p{0.09\linewidth}}
\newcolumntype{E}{>{\small\centering}p{0.115\linewidth}}
\newcolumntype{F}{>{\small\centering}p{0.18\linewidth}}
\begin{tabular} {|D|E|E|E|F|D|}
\hline Subject ID & \#Frontal Masked & \#Frontal Unmasked & \#Non-frontal Masked & \#Non-frontal Unmasked & \#Total Faces \tabularnewline
\hline
1 & 1 & 43 & 2 & 14 & 60 \tabularnewline
2 & 2 & 6 & 19 & 25 & 52 \tabularnewline
3 & 12 & 8 & 25 & 6 & 51 \tabularnewline
4 & 3 & 9 & 10 & 28 & 50 \tabularnewline
5 & 8 & 18 & 14 & 12 & 52 \tabularnewline
6 & 3 & 5 & 10 & 32 & 50 \tabularnewline
7 & 0 & 27 & 0 & 23 & 50 \tabularnewline
\hline Total & 29 & 116 & 80 & 140 & 365\tabularnewline
\hline
\end{tabular}
\end{table}

\begin{table}[!t]
\caption{A summary of probe set in terms of the \#faces with and without spectacles}
\label{t4}
\centering
\newcolumntype{D}{>{\small\centering}p{0.16\linewidth}}
\newcolumntype{E}{>{\small\centering}p{0.23\linewidth}}
\begin{tabular} {|D|E|E|D|}
\hline Subject ID & \#With Spectacles & \#Without Spectacles & \#Total Faces\tabularnewline
\hline
1 & 24 & 36 & 60 \tabularnewline
2 & 18 & 34 & 52 \tabularnewline
3 & 32 & 19 & 51 \tabularnewline
4 & 12 & 38 & 50 \tabularnewline
5 & 12 & 40 & 52 \tabularnewline
6 & 12 & 38 & 50 \tabularnewline
7 & 0 & 50 & 50 \tabularnewline
\hline Total & 110 & 255 & 365\tabularnewline
\hline
\end{tabular}
\end{table}

\begin{figure*}[!t]
\centering
  \includegraphics[width=0.97\linewidth, height=11cm]{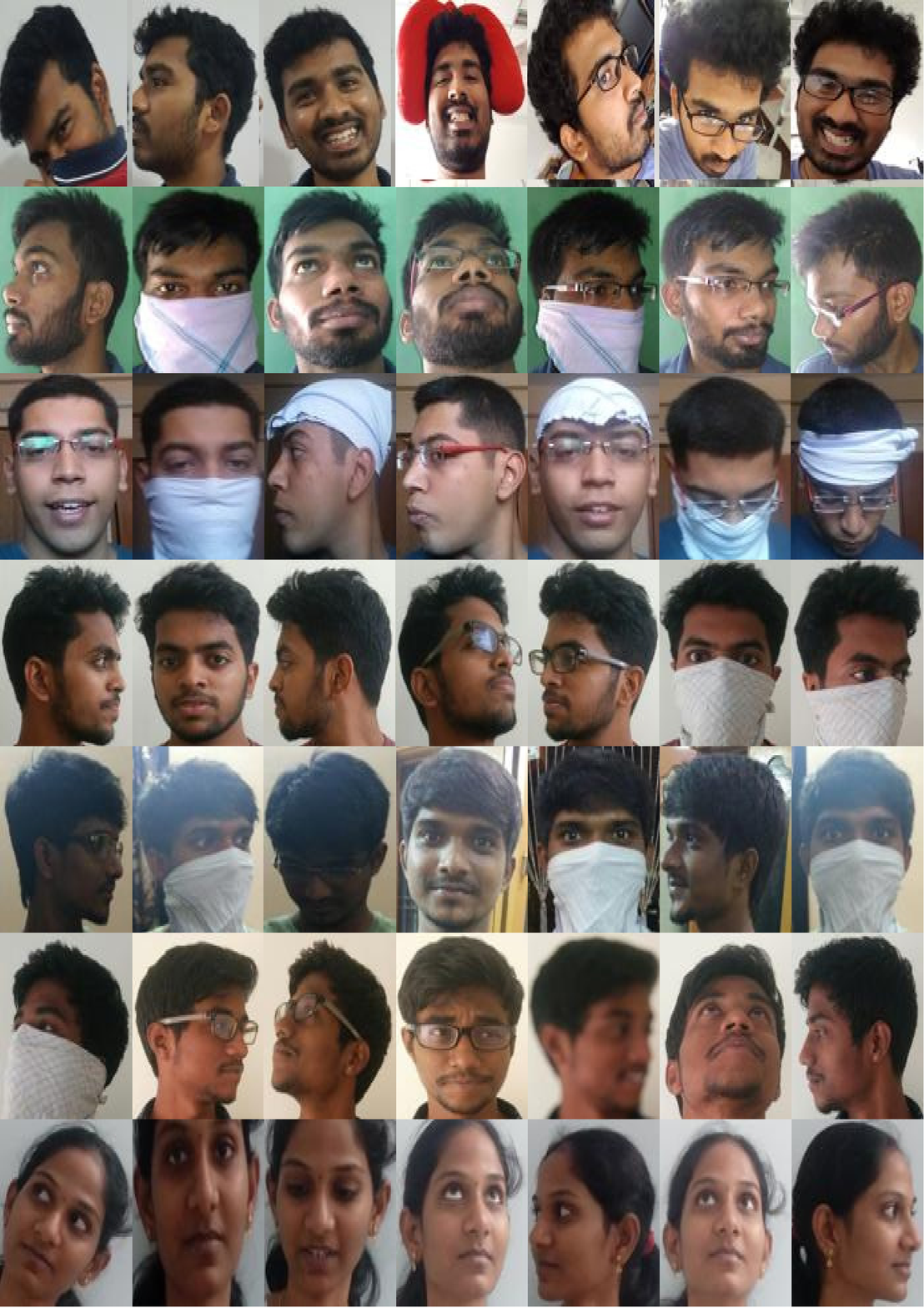}
  \caption{Sample seven faces per subject from probe set. Each row corresponds to a subject.}
  \label{fig:3}
\end{figure*}

\begin{figure*}[!t]
\centering
\includegraphics[width=1\linewidth, height=10cm]{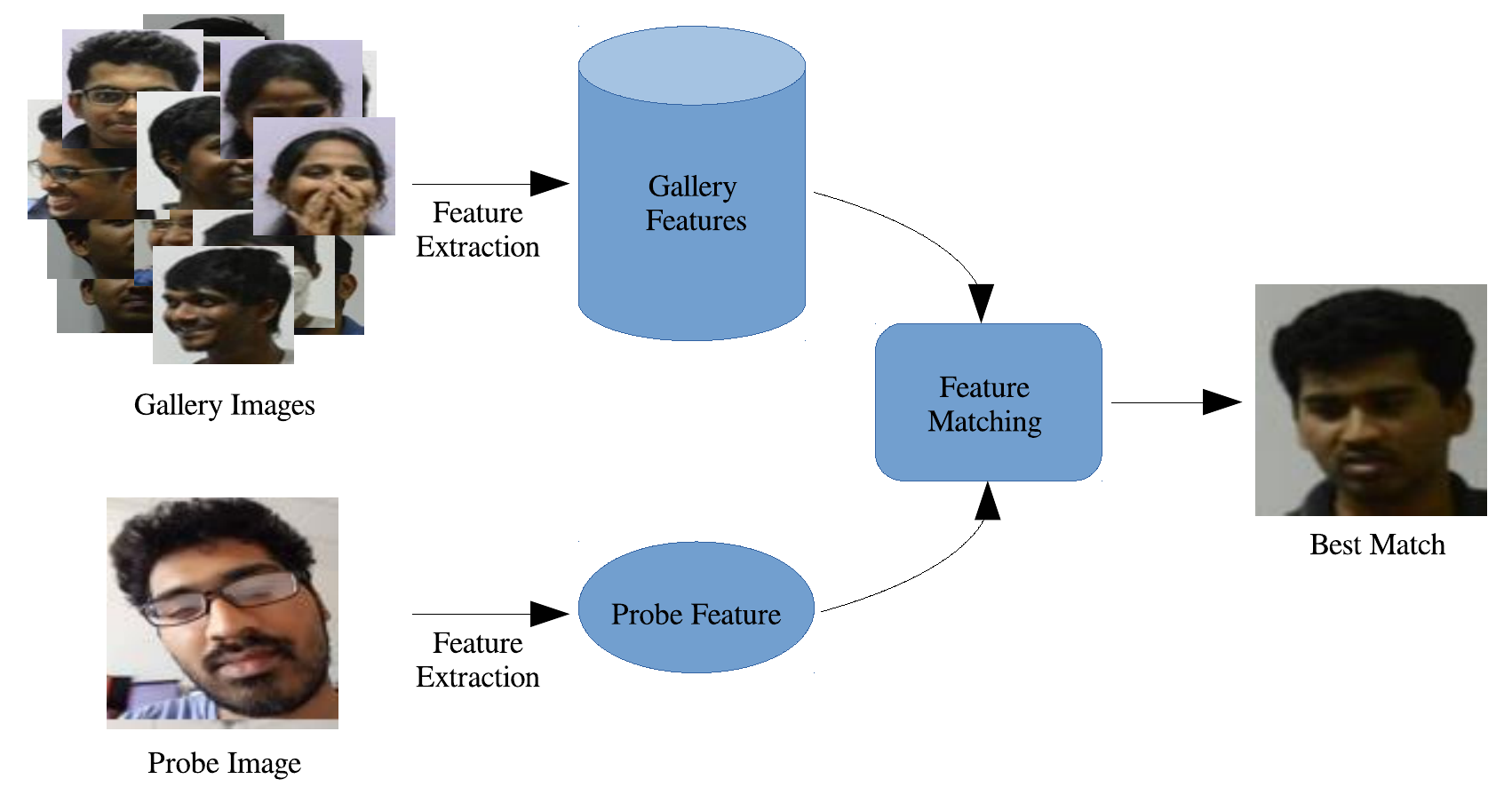}
  \caption{The face recognition framework using local descriptors. The best matching face against a probe face is extracted based on the minimum distance between feature descriptors of probe face and gallery faces.}
  \label{fig4}
\end{figure*}

\subsection{Gallery Set}
The images of the gallery set are captured from mobile phones with multiple people involved in some activities like talking, laughing, etc. A total of 180 such images are captured with minimum three and maximum five number of people in an image. Sample images of this set are shown in Figure \ref{fig:1}. We have created a cropped version of the gallery set. All the visible faces in all the images are manually cropped and annotated with the subject labels. The cropped galley set comprises of 688 faces from 180 original multi-face gallery images. The co-ordinates of each face in each image is also provided to validate a face detection algorithm. The cropped version of gallery set can be used for the experiment purpose. The characteristics of gallery set such as frontal/non-frontal pose and masked/unmasked faces are summarized in Table \ref{t1}. In gallery set, a subject is either with spectacle or without spectacle. Table \ref{t2} highlights the subjects with/without spectacles. Only subjects 3 \& 4 are with spectacles. Some cropped faces of gallery set are also shown in Figure \ref{fig:2}.

\subsection{Probe Set}
The probe set is created in the second section with same set of subjects used in galley set. For each subject, we provide a set of face images with differents poses, captured from mobile phones. Since these images are captured by the subjects individually, a lot of variations are present in the image such as occlusion, spectacle, illumination, pose, viewpoint, blur, masked, etc. Total 365 images are present in the probe set consisting of nearly 50 images from each subject. A detailed description of the probe set is illustrated in Table \ref{t3} and \ref{t4} along with the frontal/non-frontal/masked/unmasked/spectacles number of images. It can be noted that the subjects in gallery set have either used or not used the spectacles, whereas in the probe set, all the subjects except last one have mixed images with and without spectacles as depicted in Table \ref{t4}. Some example faces of probe set are also shown in Fig. \ref{fig:3} in order to illustrate the complexity of the probe set. Next we illustrate the experiments made on proposed face dataset.

\section{Face Recognition using Local Descriptors}
In this section, the nearest neighbour based face recognition framework using local descriptors is described as shown in Fig. \ref{fig4}. The features using a local descriptor is computed over gallery faces to create the gallery features database. The same descriptor is then used to extract the feature for any probe image. After computing the descriptors, the distance between probe feature and gallery features are computed. Finally, the class of probe face is recognized as the class of best matching gallery face based on the minimum distance between probe face and gallery faces. 

Several state-of-the-art face descriptors including hand-crafted and deep learned like Local Binary Pattern (LBP) \cite{lbp}, Local Ternary Pattern (LTP) \cite{ltp}, Local Derivative Pattern {LDP} \cite{ldp}, Local Directional Gradient Pattern (LDGP) \cite{ldgp}, Semi-structure Local Binary Pattern (SLBP) \cite{slbp}, Local Vector Pattern (LVP) \cite{lvp}, Local Gradient Hexa Pattern (LGHP) \cite{lghp} and VGGFace CNN descriptor \cite{vggface} are tested over the proposed dataset to establish its complexity. Note that all these descriptors are proposed for face representation purpose and VGGFace CNN descriptor is very discriminative for face representation. The MatConvNet pre-trained model of VGGFace CNN descriptor is used in this paper\footnote{http://www.vlfeat.org/matconvnet/pretrained/}. Several distances such as Euclidean, L1, Cosine, Emd (Earth Mover Distance) and Chisq (Chi-square) \cite{distance} are also used in this paper to find the best performing distance measure for the proposed dataset.

\section{Experimental Results}
The average recognition rate for the descriptors on the proposed IIITS\_MFace dataset, is used as the evaluation criteria for the descriptors. The average recognition rate is computed by taking the mean of average accuracies obtained over all the subjects of the probe set. The average accuracy for a particular subject of probe set is computed by taking the mean of accuracies obtained by turning each image of that subject as the probe image. Until and otherwise not stated, L1 distance is used to compare the descriptors.

\begin{table*}[!t]
\caption{The average recognition rate using LBP, LTP, LDP, LDGP, SLBP, LVP, LGHP and VGGFace descriptors with L1 distance over proposed IIITS\_MFace dataset.}
\label{t5}
\centering
\newcolumntype{D}{>{\small\centering}p{0.10\linewidth}}
\newcolumntype{E}{>{\small\centering}p{0.08\linewidth}}
\newcolumntype{F}{>{\small\centering}p{0.08\linewidth}}
\begin{tabular} {|D|E|E|E|F|E|F|F|E|}
\hline Descriptor & Subject1 & Subject2 & Subject3 & Subject4 & Subject5 & Subject6 & Subject7 & Mean\tabularnewline
\hline
LBP&16.67&19.23&58.82&6&19.23&20&52&27.42\tabularnewline
LTP&16.67&17.31&54.90&8&5.77&18&54&24.95\tabularnewline
LDP&21.67&30.77&64.71&0&5.77&4&2&18.42\tabularnewline
LDGP&8.33&15.38&43.14&10&1.92&8&72&22.68\tabularnewline
SLBP&11.67&23.08&45.10&4&34.62&6&88&30.35\tabularnewline
LVP&18.33&26.92&64.71&18&7.69&12&56&29.09\tabularnewline
LGHP&25&30.77&58.82&10&26.92&4&100&36.50\tabularnewline
VGGFace&83.33&51.92&68.63&32&92.31&50&100&68.31\tabularnewline
\hline
\end{tabular}
\end{table*}

\begin{table*}[!t]
\caption{Confusion matrix of average recognition rate using VGGFace descriptor with L1 distance over proposed IIITS\_MFace dataset. The True Positive Values are highlighted in bold.}
\label{t9}
\centering
\begin{tabu} to 0.95\textwidth{|X[c]|X[c]|X[c]|X[c]|X[c]|X[c]|X[c]|X[c]|X[c]|}
\hline Subjects & Subject1 & Subject2 & Subject3 & Subject4 & Subject5 & Subject6 & Subject7\\
\hline
Sub1&\textbf{50}&3&4&1&2&0&0\\
Sub2&8&\textbf{27}&2&1&12&2&0\\
Sub3&0&1&\textbf{35}&4&4&2&5\\
Sub4&9&4&20&\textbf{16}&1&0&0\\
Sub5&0&0&1&0&\textbf{48}&3&0\\
Sub6&6&1&6&0&12&\textbf{25}&0\\
Sub7&0&0&0&0&0&0&\textbf{50}\\
\hline
\end{tabu}
\end{table*}

\begin{table*}[!t]
\caption{The average recognition rate using each descriptor with different distances over proposed IIITS\_MFace dataset. The top value in a row is highlighted in bold face.}
\label{t6}
\centering
\begin{tabu} to 0.95\textwidth{|X[c]|X[c]|X[c]|X[c]|X[c]|X[c]|}
\hline Descriptor & Euclidean Distance & L1 Distance & Cosine Distance & Emd Distance & Chi-square Distance\\
\hline
LBP&25.94&\textbf{27.42}&27.38&18.78&26.66\\
LTP&22.73&24.95&22.73&22.39&\textbf{25.57}\\
LDP&19.01&18.42&20.36&\textbf{22.87}&22.85\\
LDGP&23.88&22.68&\textbf{25.48}&16.08&21.68\\
SLBP&30.27&30.35&29.99&25.96&\textbf{32.42}\\
LVP&23.67&29.09&26.27&24.07&\textbf{30.22}\\
LGHP&29.09&\textbf{36.50}&33.31&23.84&35.38\\
VGGFace&62.58&68.31&68.11&36.55&\textbf{69.39}\\
\hline
\end{tabu}
\end{table*}

\begin{table*}[!t]
\caption{A comparison of proposed IIITS\_MFace dataset with AT\&T, AR, Yale, LFW, PaSC and PubFig datasets. Here, `Y', `N' and `P' represent the presence, absence and partial presence of effects like Non-frontal (NoFront), Masked, Occlusion (Occl), Mixed-spectacle (MixSpec), Illumination variation (IllVar), Extreme-Illumination (ExtIll), Background Variation (BackVar), and MutiFace VGGFace. The last row presents the accuracy in \% using VGGFace CNN descriptor over each database using L1 distance measure.}
\label{tab}
\centering
\newcolumntype{D}{>{\small\centering}p{0.16\linewidth}}
\newcolumntype{E}{>{\small\centering}p{0.07\linewidth}}
\newcolumntype{F}{>{\small\centering}p{0.07\linewidth}}
\newcolumntype{G}{>{\small\centering}p{0.07\linewidth}}
\newcolumntype{H}{>{\small\centering}p{0.14\linewidth}}
\begin{tabular} {|D|E|F|F|F|F|G|H|}
\hline Traits & AT\&T & AR & Yale & LFW & PaSC & PubFig & IIITS\_MFace (Ours) \tabularnewline
\hline NoFront & N & N & N & Y & Y & Y & Y \tabularnewline
Masked & N & Y & N & N & N & N & Y \tabularnewline
Occl. & N & N & N & Y & N & N & Y \tabularnewline
MixSpec & Y & Y & N & Y & N & N & Y \tabularnewline
IllVar & N & N & Y & Y & Y & Y & Y \tabularnewline
ExtIll & N & N & Y & N & N & N & N \tabularnewline
BackVar & N & N & N & N & Y & Y & Y \tabularnewline
MultiFace & N & N & N & N & N & P & Y \tabularnewline
\hline 
VGGFace Result (\%) & 100 & 89.98 & 76.56 & 88.37 & 85.45 & 86.73 & 68.31  \tabularnewline
\hline
\end{tabular}
\end{table*}

The average recognition rate over proposed dataset using different descriptors with L1 distance is summarized in Table \ref{t5}. The VGGFace descriptor is the best performing one with 68.31\% average recognition rate among all the descriptors. Among hand-crafted descriptors, the LGHP is the best performing descriptor. Whereas, the LDP is the least performing descriptor because it is more suited to the frontal faces. The performance of most of the descriptor is better for Subject 7 because it is the only female subject. All the descriptors are failed to perform well in case of Subject 4 due to the following reasons: a) the number of faces in gallery set corresponding to subject 4 is just 17, b) all faces of subject 4 in the gallery set are unmasked and with spectacles, and c) the faces of subject 4 in probe set are mixed with huge amount of pose variations, with/without spectacles and masked/unmasked. Overall, despite of being recent, well-known and highly discriminative, these face descriptors are failed to perform well over the proposed face dataset.
Table \ref{t9} illustrates the confusion matrix over proposed dataset obtained using the VGGFace CNN descriptor. It can be noted that most of the Subject 4 and 6 probe faces are recognized as the Subject 1, 3 ad 5 due to the huge amount of illumination change in the probe faces of Subject 4 and 6 as compared to the gallery faces. Subjects 1, 3 and 5 are also facing the problems like illumination, background, occlusion and masking.

In order to find out which distance is better suited for the proposed IIITS\_MFace dataset, we have conducted an experiment with different distance measures such as Euclidean (Eucld), L1, Cosine, Earth Movers Distance (Emd) and Chisq (Chi-square). The average recognition rate using all the descriptors are presented in Table \ref{t6}. It can be noted that the Chi-square distance is performing well with LTP, SLBP, LVP and VGGFace descriptors. The Euclidean distance is not recommended to be used for the proposed dataset. Though, we have used L1 distance in other experiments, the best result (i.e., 69.39\% accuracy) is obtained using VGGFace descriptor using Chi-square distance.

There are challenging datasets available in the literature with challenges like different side poses, occluded faces, varying light intensities, etc. These datasets are discussed in the Introduction section. However, the proposed IIITS\_MFace dataset is much more challenging compared to the other existing face datasets such as AT\&T, AR, Yale, LFW, PaSC and PubFig, as depicted in Table \ref{tab}. The result of VGGFace descriptor is lowest over the proposed IIITS\_MFace dataset, which shows its difficulty and robustness.

From the experimental results, we can deduce that the proposed IIITS\_MFace dataset is more challenging compared to the existing face datasets even for the deep learned VGGface descriptor, which makes it more realistic for the experiments to meet the real world scenario.

\section{Conclusion}
A multi-face challenging IIITS\_MFace dataset is proposed in this paper to validate the performance of hand-crafted local descriptors as well as deep learned CNN descriptor against the different kind of variations. The difficulties like pose, illumination, occlusion, masking, spectacle, background etc. are present in the dataset. The recent state-of-the-art face image descriptors such as LBP, LGHP, VGGFace etc. are used to test the complexity of the IIITS\_MFace dataset. The results in terms of the average recognition rate support the challenges present in the dataset as the best performing VGGFace CNN descriptor achieved only 69.39\% of accuracy in best setting. In general, the VGGFace CNN descriptor is very discriminative and performs reasonably good for face recognition. Several distance measures are also tested and found that the Chi-square distance is better suited for this dataset. In future, the number of subjects and number of samples in the dataset may be increased to facilitate applying some deeper neural network architecture for more robust training.

\section*{Acknowledgment}
The authors would like to thank all the individuals who have been involved in the process of data collection. Special thanks to Kanv Kumar and Naveen Thella for capturing the images.

\bibliographystyle{IEEEtran}
\bibliography{reference}

\begin{thebibliography}{10}
\providecommand{\url}[1]{#1}
\csname url@samestyle\endcsname
\providecommand{\newblock}{\relax}
\providecommand{\bibinfo}[2]{#2}
\providecommand{\BIBentrySTDinterwordspacing}{\spaceskip=0pt\relax}
\providecommand{\BIBentryALTinterwordstretchfactor}{4}
\providecommand{\BIBentryALTinterwordspacing}{\spaceskip=\fontdimen2\font plus
\BIBentryALTinterwordstretchfactor\fontdimen3\font minus
  \fontdimen4\font\relax}
\providecommand{\BIBforeignlanguage}[2]{{%
\expandafter\ifx\csname l@#1\endcsname\relax
\typeout{** WARNING: IEEEtran.bst: No hyphenation pattern has been}%
\typeout{** loaded for the language `#1'. Using the pattern for}%
\typeout{** the default language instead.}%
\else
\language=\csname l@#1\endcsname
\fi
#2}}
\providecommand{\BIBdecl}{\relax}
\BIBdecl

\bibitem{pietikainen}
M.~Pietik{\"a}inen, A.~Hadid, G.~Zhao, and T.~Ahonen, ``Face analysis using
  still images,'' in \emph{Computer Vision Using Local Binary Patterns}.\hskip
  1em plus 0.5em minus 0.4em\relax Springer, 2011, pp. 151--168.

\bibitem{bereta}
M.~Bereta, W.~Pedrycz, and M.~Reformat, ``Local descriptors and similarity
  measures for frontal face recognition: a comparative analysis,''
  \emph{Journal of Visual Communication and Image Representation}, vol.~24,
  no.~8, pp. 1213--1231, 2013.

\bibitem{smeulders}
A.~W. Smeulders, M.~Worring, S.~Santini, A.~Gupta, and R.~Jain, ``Content-based
  image retrieval at the end of the early years,'' \emph{IEEE Transactions on
  pattern analysis and machine intelligence}, vol.~22, no.~12, pp. 1349--1380,
  2000.

\bibitem{huang}
D.~Huang, C.~Shan, M.~Ardabilian, Y.~Wang, and L.~Chen, ``Local binary patterns
  and its application to facial image analysis: a survey,'' \emph{IEEE
  Transactions on Systems, Man, and Cybernetics, Part C (Applications and
  Reviews)}, vol.~41, no.~6, pp. 765--781, 2011.

\bibitem{yang}
B.~Yang and S.~Chen, ``A comparative study on local binary pattern (lbp) based
  face recognition: Lbp histogram versus lbp image,'' \emph{Neurocomputing},
  vol. 120, pp. 365--379, 2013.

\bibitem{lfw}
E.~Learned-Miller, G.~B. Huang, A.~RoyChowdhury, H.~Li, and G.~Hua, ``Labeled
  faces in the wild: A survey,'' in \emph{Advances in face detection and facial
  image analysis}.\hskip 1em plus 0.5em minus 0.4em\relax Springer, 2016, pp.
  189--248.

\bibitem{attdb}
``\uppercase{AT\&T} face database,''
  \url{http://www.cl.cam.ac.uk/research/dtg/attarchive/-facedatabase.html}.

\bibitem{ardb}
``\uppercase{AR} face database,''
  \url{http://www2.ece.ohio-state.edu/~aleix/ARdatabase.html}.

\bibitem{CroppedYale}
K.~Lee, J.~Ho, and D.~Kriegman, ``Acquiring linear subspaces for face
  recognition under variable lighting,'' \emph{IEEE Trans. Pattern Anal. Mach.
  Intelligence}, vol.~27, no.~5, pp. 684--698, 2005.

\bibitem{lfwdb}
``\uppercase{LFW} face database,'' \url{http://vis-www.cs.umass.edu/lfw/}.

\bibitem{CroppedLFW}
``Croppedlfw face database,'' \url{http://conradsanderson.id.au/lfwcrop/}.

\bibitem{pasc}
J.~R. Beveridge, P.~J. Phillips, D.~S. Bolme, B.~A. Draper, G.~H. Givens, Y.~M.
  Lui, M.~N. Teli, H.~Zhang, W.~T. Scruggs, K.~W. Bowyer \emph{et~al.}, ``The
  challenge of face recognition from digital point-and-shoot cameras,'' in
  \emph{Biometrics: Theory, Applications and Systems (BTAS), 2013 IEEE Sixth
  International Conference on}.\hskip 1em plus 0.5em minus 0.4em\relax IEEE,
  2013, pp. 1--8.

\bibitem{viola}
P.~Viola and M.~Jones, ``Rapid object detection using a boosted cascade of
  simple features,'' in \emph{Computer Vision and Pattern Recognition, 2001.
  CVPR 2001. Proceedings of the 2001 IEEE Computer Society Conference on},
  vol.~1.\hskip 1em plus 0.5em minus 0.4em\relax IEEE, 2001, pp. I--I.

\bibitem{pubfig}
N.~Kumar, A.~C. Berg, P.~N. Belhumeur, and S.~K. Nayar, ``Attribute and simile
  classifiers for face verification,'' in \emph{Computer Vision, 2009 IEEE 12th
  International Conference on}.\hskip 1em plus 0.5em minus 0.4em\relax IEEE,
  2009, pp. 365--372.

\bibitem{lbp}
T.~Ahonen, A.~Hadid, and M.~Pietikainen, ``Face description with local binary
  patterns: Application to face recognition,'' \emph{IEEE transactions on
  pattern analysis and machine intelligence}, vol.~28, no.~12, pp. 2037--2041,
  2006.

\bibitem{ltp}
X.~Tan and B.~Triggs, ``Enhanced local texture feature sets for face
  recognition under difficult lighting conditions,'' \emph{IEEE transactions on
  image processing}, vol.~19, no.~6, pp. 1635--1650, 2010.

\bibitem{ldp}
B.~Zhang, Y.~Gao, S.~Zhao, and J.~Liu, ``Local derivative pattern versus local
  binary pattern: face recognition with high-order local pattern descriptor,''
  \emph{IEEE transactions on image processing}, vol.~19, no.~2, pp. 533--544,
  2010.

\bibitem{ldgp}
S.~Chakraborty, S.~K. Singh, and P.~Chakraborty, ``Local directional gradient
  pattern: a local descriptor for face recognition,'' \emph{Multimedia Tools
  and Applications}, vol.~76, no.~1, pp. 1201--1216, 2017.

\bibitem{ldop}
S.~R. Dubey and S.~Mukherjee, ``Ldop: Local directional order pattern for
  robust face retrieval,'' \emph{arXiv preprint arXiv:1803.07441}, 2018.

\bibitem{slbp}
K.~Jeong, J.~Choi, and G.-J. Jang, ``Semi-local structure patterns for robust
  face detection,'' \emph{IEEE Signal Processing Letters}, vol.~22, no.~9, pp.
  1400--1403, 2015.

\bibitem{lvp}
K.-C. Fan and T.-Y. Hung, ``A novel local pattern descriptor—local vector
  pattern in high-order derivative space for face recognition,'' \emph{IEEE
  transactions on image processing}, vol.~23, no.~7, pp. 2877--2891, 2014.

\bibitem{lghp}
S.~Chakraborty, S.~Singh, and P.~Chakraborty, ``Local gradient hexa pattern: A
  descriptor for face recognition and retrieval,'' \emph{IEEE Transactions on
  Circuits and Systems for Video Technology}, 2016.

\bibitem{vggface}
O.~M. Parkhi, A.~Vedaldi, A.~Zisserman \emph{et~al.}, ``Deep face
  recognition.'' in \emph{BMVC}, vol.~1, no.~3, 2015, p.~6.

\bibitem{distance}
``Pairwise distance between two sets of observations,''
  \url{http://in.mathworks.com/help/stats/pdist2.html}.

\end{thebibliography}

\end{document}